\documentclass{article}
\usepackage{ijcai24}

\usepackage{times}
\usepackage{soul}
\usepackage{url}
\usepackage[hidelinks]{hyperref}
\usepackage[utf8]{inputenc}
\usepackage[small]{caption}
\usepackage{graphicx}
\usepackage{amsmath}
\usepackage{amsthm}
\usepackage{amsfonts}
\usepackage{booktabs}
\usepackage{enumitem}
\usepackage{algorithm}
\usepackage{algorithmic}
\usepackage[switch]{lineno}

\usepackage{mydef}

\newcommand{\paratitle}[1]{\vspace{0.5ex}\noindent\textbf{\textit{#1}}}

\def \Y {\mathbf{Y}}

\title{A Survey of Data-Efficient Graph Learning}

\author{
Wei Ju$^1$\and
Siyu Yi$^{2*}$\and
Yifan Wang$^3$\and
Qingqing Long$^1$\and
Junyu Luo$^1$\and
Zhiping Xiao$^4$\and
Ming Zhang$^{1}$\thanks{Corresponding authors.}
\affiliations
$^1$School of Computer Science, Peking University\and \\
$^2$School of Statistics and Data Science, Nankai University\and \\
$^3$School of Information Technology $\&$ Management, University of International Business and Economics\and \\
$^4$Department of Computer Science, University of California Los Angeles
\emails
\{juwei, qingqinglong, mzhang\_cs\}@pku.edu.cn,
siyuyi@mail.nankai.edu.cn, yifanwang@uibe.edu.cn,  luojunyu@stu.pku.edu.cn, patricia.xiao@cs.ucla.edu
}

\begin{document}

\maketitle

\begin{abstract}
Graph-structured data, prevalent in domains ranging from social networks to biochemical analysis, serve as the foundation for diverse real-world systems. While graph neural networks demonstrate proficiency in modeling this type of data, their success is often reliant on significant amounts of labeled data, posing a challenge in practical scenarios with limited annotation resources. To tackle this problem, tremendous efforts have been devoted to enhancing graph machine learning performance under low-resource settings by exploring various approaches to minimal supervision. In this paper, we introduce a novel concept of Data-Efficient Graph Learning (DEGL) as a research frontier, and present the first survey that summarizes the current progress of DEGL. We initiate by highlighting the challenges inherent in training models with large labeled data, paving the way for our exploration into DEGL. Next, we systematically review recent advances on this topic from several key aspects, including self-supervised graph learning, semi-supervised graph learning, and few-shot graph learning. Also, we state promising directions for future research, contributing to the evolution of graph machine learning.
\end{abstract}

\section{Introduction}

Graph learning has emerged as a pivotal field at the intersection of machine learning and graph theory, offering a versatile framework for modeling and analyzing complex relationships in various domains~\cite{jin2020self,ju2024comprehensive}. With the increasing prevalence of graph-structured data, ranging from social networks to biological interactions~\cite{sun2020infograph}, the demand for effective methods to extract meaningful insights from graphs has grown substantially.

With the rising popularity of graph neural networks, particularly notable for their effective message-passing mechanisms~\cite{kipf2017semi}, a myriad of graph-related challenges have witnessed outstanding performance. However, despite the promising advancements in graph learning, the current landscape predominantly relies on extensive labeled data, introducing challenges associated with high annotation costs, time-intensive processes, and resource demands~\cite{hao2020asgn,luo2023towards}. This limitation becomes particularly evident in practical applications where obtaining substantial labeled data is impractical. To illustrate this imperative, 
in molecular domain, annotating intricate molecular structures or interactions could involve sophisticated experiments and the engagement of specialized scientists, leading to elevated annotation costs~\cite{ju2023few}; in the genomics research, acquiring precisely annotated gene functions or interactions might demand the expertise and time of professional biologists for understanding complex relationships~\cite{yu2023topological}.
These instances highlight the urgent need for graph learning methodologies tailored to low-resource settings, emphasizing the critical necessity of developing approaches that can operate effectively with limited labeled data.

\tikzstyle{leaf}=[draw=black, 
    rounded corners,minimum height=1em,
    text width=24.50em, edge=black!10, 
    text opacity=1, align=center,
    fill opacity=.3,  text=black,font=\scriptsize,
    inner xsep=3pt, inner ysep=1pt,
    ]
\tikzstyle{leaf1}=[draw=black, 
    rounded corners,minimum height=1em,
    text width=6.28em, edge=black!10, 
    text opacity=1, align=center,
    fill opacity=.5,  text=black,font=\scriptsize,
    inner xsep=3pt, inner ysep=1pt,
    ]
\tikzstyle{leaf2}=[draw=black, 
    rounded corners,minimum height=1em,
    text width=6.28em, edge=black!10, 
    text opacity=1, align=center,
    fill opacity=.8,  text=black,font=\scriptsize,
    inner xsep=3pt, inner ysep=1pt,
    ]
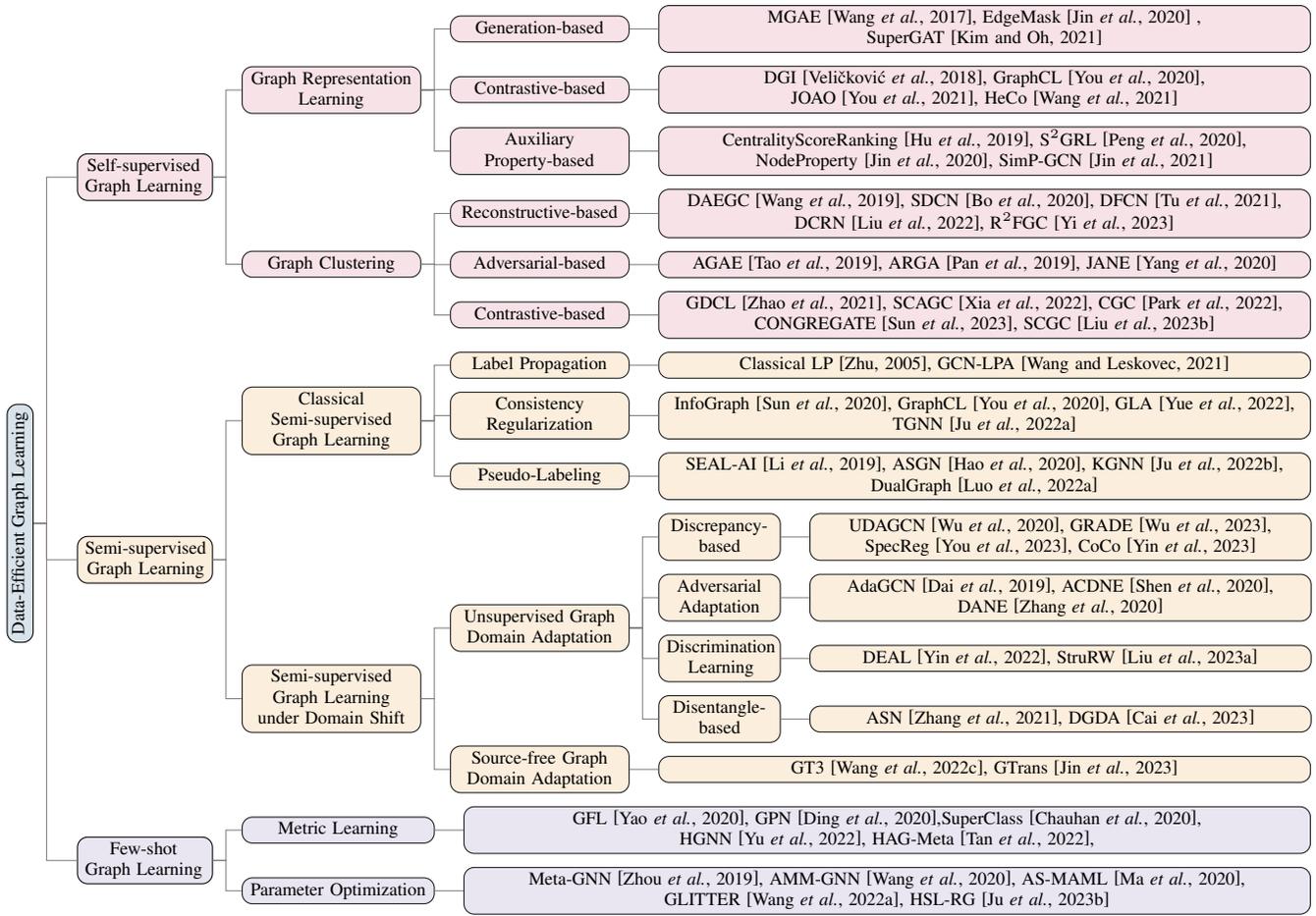
\begin{figure*}[ht]
\centering
\begin{forest}
  for tree={
  forked edges,
  grow=east,
  reversed=true,
  anchor=base west,
  parent anchor=east,
  child anchor=west,
  base=middle,
  font=\scriptsize,
  rectangle,
  draw=black, 
  edge=black!50, 
  rounded corners,
  align=center,
  minimum width=2em,
  s sep=5pt,
  inner xsep=3pt,
  inner ysep=1pt
  },
  where level=1{text width=4.5em}{},
  where level=2{text width=6em,font=\scriptsize}{},
  where level=3{font=\scriptsize}{},
  where level=4{font=\scriptsize}{},
  where level=5{font=\scriptsize}{},
  [Data-Efficient Graph Learning,rotate=90,anchor=north,edge=black!50,fill=myblue,draw=black
    [Self-supervised \\ Graph Learning,edge=black!50,text width=4.6em, fill=myred
        [Graph Representation \\Learning, leaf2, fill=myred
            [Generation-based, leaf1,fill=myred
                 [MGAE \cite{wang2017mgae}{,} EdgeMask \cite{jin2020self} {,} \\SuperGAT \cite{kim2021find},leaf,fill=myred]
            ]
            [Contrastive-based, leaf1,fill=myred
                 [DGI \cite{velivckovic2018deep}{,} GraphCL \cite{you2020graph}{,} \\JOAO \cite{you2021graph}{,} HeCo \cite{wang2021self}, ,leaf,fill=myred]
            ]
            [Auxiliary \\ Property-based, leaf1,fill=myred
                 [CentralityScoreRanking \cite{hu2019pre}{,}
                 S$^2$GRL \cite{peng2020self}{,}  \\NodeProperty \cite{jin2020self}{,} SimP-GCN \cite{jin2021node},leaf,fill=myred]
            ] 
        ]
        [Graph Clustering, leaf2,fill=myred
            [Reconstructive-based, leaf1, fill=myred
                 [DAEGC \cite{wang2019attributed}{,} SDCN \cite{bo2020structural}{,} DFCN \cite{tu2021deep}{,}\\ DCRN \cite{liu2022deep}{,} R$^2$FGC \cite{yi2023redundancy},leaf, edge=black!50, fill=myred]
            ] 
            [Adversarial-based, leaf1, fill=myred
                 [AGAE \cite{tao2019adversarial}{,} ARGA \cite{pan2019learning}{,} JANE \cite{yang2020jane},leaf,fill=myred]
            ] 
            [Contrastive-based, leaf1, fill=myred
                 [GDCL \cite{zhao2021graph}{,} SCAGC \cite{xia2022self}{,} CGC \cite{park2022cgc}{,} \\ CONGREGATE \cite{sun2023congregate}{,} SCGC \cite{liu2023simple},leaf,fill=myred]
            ]  
        ]
    ]
    [Semi-supervised \\ Graph Learning,edge=black!50,text width=4.6em, fill=myyellow
    	[Classical \\ Semi-supervised \\ Graph Learning,leaf2, fill=myyellow,
            [Label Propagation, leaf1, fill=myyellow
                 [Classical LP~\cite{zhu2005semi}{,} GCN-LPA~\cite{wang2021combining},leaf,fill=myyellow]
            ]  
            [Consistency \\ Regularization, leaf1, fill=myyellow
                 [InfoGraph~\cite{sun2020infograph}{,} GraphCL~\cite{you2020graph}{,} GLA~\cite{yue2022label}{,} \\ TGNN~\cite{ju2022tgnn},leaf,fill=myyellow]
            ]  
            [Pseudo-Labeling, leaf1, fill=myyellow
                 [SEAL-AI~\cite{li2019semi}{,} ASGN~\cite{hao2020asgn}{,} KGNN~\cite{ju2022kgnn}{,} \\ DualGraph~\cite{luo2022dualgraph},leaf,fill=myyellow]
            ]  
         ]
         [Semi-supervised \\ Graph Learning \\ under Domain Shift, leaf2, fill=myyellow
            [Unsupervised Graph \\ Domain Adaptation, leaf1, fill=myyellow
                 [Discrepancy- \\ based,leaf1,text width=4.08em,fill=myyellow
                     [UDAGCN~\cite{udagcn}{,} GRADE~\cite{grade}{,} \\ SpecReg~\cite{SpecReg}{,} CoCo~\cite{coco},leaf,text width=18.70em,fill=myyellow]
                 ]
                 [Adversarial \\ Adaptation,leaf1,text width=4.08em,fill=myyellow
                     [AdaGCN~\cite{adagcn}{,} ACDNE~\cite{acdne}{,} \\ DANE~\cite{dane},leaf,text width=18.70em,fill=myyellow]
                 ]
                 [Discrimination \\ Learning,leaf1,text width=4.08em,fill=myyellow
                     [DEAL~\cite{deal}{,} StruRW~\cite{liu2023structural},leaf,text width=18.70em,fill=myyellow]
                 ]
                 [Disentangle-\\ based,leaf1,text width=4.08em,fill=myyellow
                     [ASN~\cite{asn}{,} DGDA~\cite{DGDA},leaf,text width=18.70em,fill=myyellow]
                 ]
            ] 
            [Source-free Graph \\ Domain Adaptation, leaf1, fill=myyellow
                 [GT3~\cite{gt3}{,} GTrans~\cite{jin2023empowering}, leaf,fill=myyellow]
            ]
         ]
    ]
    [Few-shot \\ Graph Learning,edge=black!50,text width=4.6em, fill=mypurple
    	[Metric Learning,leaf1,text width=6.8em,fill=mypurple
	    [GFL~\cite{yao2020graph}{,} GPN \cite{ding2020graph}{,}SuperClass~\cite{chauhan2020few}{,} \\ HGNN~\cite{yu2022hybrid}{,} HAG-Meta \cite{tan2022graph}{,},leaf,text width=32.00em, fill=mypurple]
            ]
	[Parameter Optimization,leaf1,text width=6.8em,fill=mypurple
	    [Meta-GNN~\cite{zhou2019meta}{,} AMM-GNN~\cite{wang2020graph}{,}  AS-MAML~\cite{ma2020adaptive}{,}\\  GLITTER~\cite{wang2022graph}{,} HSL-RG~\cite{ju2023few},leaf,text width=32.00em, fill=mypurple]
	]
    ]
  ]
\end{forest}
\caption{A taxonomy of data-efficient graph learning (DEGL).}
\label{fig:taxonomy_of_DEGL}
\end{figure*}

For effective graph learning, researchers have conducted extensive and specialized studies focusing on the exploration of graph machine learning under low-resource settings, aiming at mitigating the costs and time associated with annotation. Nevertheless, although there have been increasingly applied in effective graph learning, this rapidly expanding field still lacks a systematic review. To fill this gap, in this paper, we develop a novel concept of Data-Efficient Graph Learning (DEGL) to summarize the existing works of DEGL and provide a promising research frontier to aid researchers in reviewing, summarizing, and strategizing for the future.



\section{Taxonomy}

To enhance our understanding of the dynamic evolution in DEGL, we identify pivotal research endeavors, 
analyze their motivations, and succinctly encapsulate their primary technical contributions.
As illustrated in Figure~\ref{fig:taxonomy_of_DEGL}, this survey establishes a new taxonomy, which divides these works into three different categories, i.e., self-supervised graph learning, semi-supervised graph learning, and few-shot graph learning. These groups can be further summarized in different scenarios. Then, we briefly introduce these three categories as follows:
\begin{itemize}[leftmargin=*]
\item \paratitle{Self-supervised Graph Learning} is a paradigm that leverages the inherent structure and relationships within graph data to train models without relying on external labeled information. At its core, the key idea is to design tasks that encourage the model to learn meaningful representations from the graph data itself. Based on the presence of a specific downstream task, self-supervised graph learning can be further categorized into non-end-to-end graph representation learning~\cite{jin2020self} and end-to-end graph clustering~\cite{bo2020structural,liu2022deep}. 

\item \paratitle{Semi-supervised Graph Learning} involves using both labeled and unlabeled samples to train models, capitalizing on the available information while handling scenarios with limited labeled samples. The fundamental idea is to leverage the relationships within the labeled and unlabeled instances to guide the model in learning representations for unlabeled nodes or graphs. Depending on whether there is a distribution shift during inference, we further categorize semi-supervised graph learning into classical semi-supervised graph learning~\cite{wang2021combining,ju2022tgnn} and semi-supervised graph learning under domain shift~\cite{coco,jin2023empowering}.

\item \paratitle{Few-shot Graph Learning} is a specialized area designed to enable models to effectively generalize and make accurate predictions when presented with only a limited number of labeled examples. The core idea is to equip the model with the ability to learn from a few annotated instances and then apply this acquired knowledge to make predictions on new, unseen data. Thus researchers either adopt metric learning to encourage each query node to approach its 
prototypes~\cite{tan2022graph} or parameter optimization to generate node representations using meta-learning~\cite{ju2023few}.
\end{itemize}

The three research directions are interconnected rather than mutually exclusive, contributing collectively to the rapid advancement of the data-efficient graph learning field. Table~\ref{tab:representative_works} further analyzes the representative DEGL works according to the different properties. In the following sections, we delve deeper into these three research directions, exploring key challenges, representative solutions, and emerging trends.

\begin{table}[t]
\tabcolsep=3pt
\centering
{\small
\begin{tabular}{c|c|c|c|c}
\toprule
\multicolumn{2}{c|}{Branch} & Methods & Data Type & Objective Function \\
\midrule
\multirow{8}{*}{Self} & \multirow{3}{*}{GRL} & MGAE & Node & Reconstruction Loss \\
 \cmidrule{3-5}
 & & GraphCL & Graph & InfoNCE Loss \\
  \cmidrule{3-5}
  & & S$^{2}$GRL & Graph & Reconstruction Loss \\
 \cmidrule{2-5}
 & \multirow{6}{*}{GC} & \multirow{2}{*}{DFCN} & \multirow{2}{*}{Node} & KL Loss, \\
  &  &  &  & Reconstruction Loss \\
  \cmidrule{3-5}
 & & \multirow{2}{*}{ARGA} & \multirow{2}{*}{Node} & Adversarial Loss, \\
 & & & & Reconstruction Loss \\
 \cmidrule{3-5}
  & & GLCC & Graph & InfoNCE Loss \\
\midrule
\multirow{12}{*}{Semi} & \multirow{6}{*}{CSemi} & LP & Node & Cross-Entropy Loss \\
 \cmidrule{3-5}
 &  & \multirow{2}{*}{TGNN} & \multirow{2}{*}{Graph} & Consistency Loss, \\
  &  &  &  & Cross-Entropy Loss \\
 \cmidrule{3-5}
  &  & \multirow{2}{*}{DualGraph} & \multirow{2}{*}{Graph} & InfoNCE Loss, \\
  &  &  &  & Cross-Entropy Loss \\
 \cmidrule{2-5}
 & \multirow{5}{*}{Semi w DS} & UDAGCN  & {Node}  & Cross-Entropy Loss  \\
  \cmidrule{3-5}
  & & {GTrans}  & {Node}  & Surrogate Loss \\
  \cmidrule{3-5}
   & & \multirow{2}{*}{DEAL}  & \multirow{2}{*}{Graph}  & Adversarial Loss,  \\
  & & & & Cross-Entropy Loss \\
\midrule
\multicolumn{2}{c|}{\multirow{6}{*}{Few-shot}} & \multirow{2}{*}{GFL} & \multirow{2}{*}{Node} & Reconstruction Loss \\
\multicolumn{2}{c|}{} &  &  & Cross-Entropy Loss \\
\cmidrule{3-5}
\multicolumn{2}{c|}{} &SuperClass & Graph & Cross-Entropy Loss \\
\cmidrule{3-5}
\multicolumn{2}{c|}{} & \multirow{2}{*}{HSL-RG} & \multirow{2}{*}{Graph} & InfoNCE Loss \\
\multicolumn{2}{c|}{} &  &  & Cross-Entropy Loss \\
\bottomrule
\end{tabular}
}
\caption{{\small Analysis for representative DEGL works according to the data type and objective function. Self-supervised Graph Learning (Self); Semi-supervised Graph Learning (Semi); Few-shot Graph Learning (Few-shot); Graph Representation Learning (GRL); Graph Clustering (GC); Classical Semi-supervised Graph Learning (CSemi); Semi-supervised Graph Learning under Domain Shift (Semi w DS).}}
\label{tab:representative_works}
\end{table}

\section{Self-supervised Graph Learning}

\subsection{Graph Representation Learning}
Graph Representation Learning (GRL) \cite{jin2020self} allows the model to capture nuanced patterns and dependencies among nodes without relying on manual labels, making it particularly valuable in scenarios where labels are expensive or hard to acquire. In general, the framework of GRL can be summarized as:
\begin{equation}
    \min_{\theta,\psi}\mathcal{L}\left (\mathcal{D}, f_\theta, g_\psi\right ),
    \label{eq:grl}
\end{equation}
where $\mathcal{D}$ denotes the data distribution of the unlabeled graph, $f_\theta$ denotes the node or graph encoder, aiming at learning a low-dimensional representation $h_i\in\mathcal{H}$ for node $v_i$.
The decoder $g_\theta $ takes the node representation $\mathcal{H}$ as its input and is generally well-designed on specific downstream tasks. 

Existing GRL methods can be roughly divided into three categories, i.e., generation-based, contrastive-based, and auxiliary property-based methods.

\subsubsection{Generation-based GRL}
Generation-based GRL \cite{kim2021find,wang2022disencite} aims at reconstructing the input full
graph or sampled subgraphs. These methods contribute to understanding the underlying structure and node dependencies within the graph, which encourages the model to encode representations that preserve the inherent information. The self-supervised loss is generally designed to quantify the difference between the reconstructed and the original graph. In such case, Eq.~\eqref{eq:grl} is derived as:
\begin{equation}
    \min_{\theta,\psi}\mathcal{L}\left ( g_\psi (f_\theta(\hat{\mathcal{G}})), \mathcal{G}\right ),
\end{equation}
where $\mathcal{G}$ denotes the perturbed graph data, $f_\theta(\cdot)$ and $g_\psi (\cdot)$ denote graph encoder and decoder respectively. 
MGAE \cite{wang2017mgae}, estimates raw features from the noisy input node features in each GNN layer, i.e., $\hat{\mathcal{G}}=(A,\hat{X})$,  where $\hat{X}$ denotes the corrupted random noise.
EdgeMask \cite{jin2020self} and SuperGAT \cite{jin2020self} are representative works that recover the graph structure.
Specifically, EdgeMask recovers the similarity between the embeddings of two connected nodes, and SuperGAT adopts a hierarchical variational inference strategy to reconstruct the global topological structure.

\subsubsection{Contrastive-based GRL}
Contrastive-based GRL \cite{you2020graph,luo2023self,ju2024towards} mainly incorporates contrastive learning on learning node representations by contrasting similar instances (positive pairs) against dissimilar instances (negative pairs). Contrastive-based models provide a powerful mechanism for exploiting the inherent structure and relationships within the graph. For this objective, Eq.~\eqref{eq:grl} can be derived as: 
\begin{equation}
\min_{\theta,\psi}\mathcal{L}\left ( g_\psi (f_\theta(\hat{\mathcal{G}}_{\text{view1}}), f_\theta(\hat{\mathcal{G}}_{\text{view2}}))\right ),
\end{equation}
where $\hat{\mathcal{G}}_{\text{view1}}$ and $\hat{\mathcal{G}}_{\text{view2}}$ denote the augmented instances, which can be node pairs, graph pairs, and task pairs. $g_{\psi}$ denotes the similarity discriminator, which estimates the distances of contrastive instances. GraphCL \cite{you2020graph} contrasts two generated views of node feature masking and edge reconstruction, and then the Mutal Information is maximized between two target nodes from different views. JOAO \cite{you2021graph} proposes a bi-level optimization problem, which optimizes the augmentation selection module and contrastive objectives jointly. DGI \cite{velivckovic2018deep} contrasts node-level representations with graph-level ones for better local-global interactions. For heterogeneous graphs, HeCo \cite{wang2021self} is designed to contrast two generated views of network schema and meta-paths. 

\subsubsection{Auxiliary Property-based GRL}
Auxiliary property-based GRL~\cite{zhu2020cagnn,luo2022clear} generates pseudo labels by leveraging classical attributive and graph algorithms, to enrich the self-supervision signals. Compared with the human-annotated labels in supervised learning, the pseudo labels in auxiliary property-based GRL do not need additional cost. 
The utilization of auxiliary properties in GRL enhances the robustness and generalization capability of the representations. Under this setting, Eq.~\eqref{eq:grl} can be derived as: 
\begin{equation}
  \min_{\theta,\psi}\mathcal{L}\left ( g_\psi (f_\theta(\mathcal{G})), c(\mathcal{G})\right ),
\end{equation}
where $c(\mathcal{G})$ denotes the annotated auxiliary properties, such as the shortest path and graph centrality. S$^2$GRL \cite{peng2020self} is designed to estimate the shortest paths among different nodes for better structural representations. CentralityScoreRanking \cite{hu2019pre} predicts the ranking of centrality scores among node pairs, and compares the ranking with pseudo labels. NodeProperty \cite{jin2020self} adopts a node-level property construction task aiming at exploring the properties of node features. SimP-GCN \cite{jin2021node} designs a $k$-nearest-neighbor graph based on the node features to enhance the neighbor aggregation performance. 

\subsubsection{Discussion}
Generation-based GRL is easier to implement as the reconstruction task is easy to build. But it is memory-consuming in certain cases such as dealing with large-scale graphs. 
Auxiliary property-based methods benefit from the uncomplicated decoders, while the selection of effective auxiliary properties is challenging. Compared with the above categories, contrastive-based ones are more flexible.

\subsection{Graph Clustering}

In self-supervised graph learning, an end-to-end learning paradigm that possesses a specific downstream task mainly boils down to graph clustering. Graph clustering aims to divide the nodes into $C$ disjoint clusters without label signals as reliable guidance. Recently, the graph clustering task has attracted widespread attention and extensive approaches have focused on this task with promising performance. In general, given an attribute graph $\mathcal{G}$, the framework of graph clustering includes a self-supervised neural network $F$ that outputs the node representations and a clustering network/traditional clustering $\Omega$ that exports the clustering result $\hat{\Y}$, i.e., 
$$\hat{\Y} = \Omega(F(\mathcal{G}), C).$$ 
Existing methods can be roughly partitioned into three main groups: reconstructive-based, adversarial-based, and contrastive-based approaches. In the following, we present a comprehensive overview of these algorithms.

\subsubsection{Reconstructive-based Methods}

Reconstructive-based methods typically encode either the attribute information or the structural information of the graph, aiming to reconstruct the input information to supervise the network training and achieve meaningful node representations. 
DAEGC \cite{wang2019attributed} performs representation learning through an attention network based on node features and structural information under the GAT paradigm. The network training is supervised by reconstructing the graph structure and a self-supervised clustering module. 
SDCN \cite{bo2020structural} leverages auto-encoder (AE) to assist GCN, effectively alleviating the over-smoothing issue. It obtains semantically rich node representations by optimizing the reconstruction loss of the AE and a self-supervised clustering loss based on GCN. 
DFCN \cite{tu2021deep} introduces a dynamic information fusion technique based on AE and graph auto-encoder (GAE) to explore attributive and structural information finely. It also designs a triplet self-supervision mechanism for self-supervised clustering. 
Based on \cite{tu2021deep}, DCRN \cite{liu2022deep} reduces information correlation through a dual approach to prevent representation collapse and obtain discriminative node representations. 
R$^2$FGC \cite{yi2023redundancy} builds upon the relational learning at the attributive and structural levels from both global and local views based on AE and GAE. It preserves relationships among positive samples and reduces redundant relationships among negative samples, thereby acquiring effective and discriminative node representations.

\subsubsection{Adversarial-based Methods}

Adversarial-based methods engage in a game between the generator and the discriminator to achieve robust node representations. 
AGAE \cite{tao2019adversarial} combines adversarial learning and auto-encoder to perform representation learning, which introduces an adversarial regularization term and adaptive partition-dependent prior to guide the network training. 
ARGA \cite{pan2019learning} incorporates an adversarial training scheme into the graph auto-encoder architecture to regularize the latent codes for learning a robust graph representation. The adversarial training module is designed to discern whether the latent codes originate from a real prior distribution or the graph encoder.
Further, JANE \cite{yang2020jane} designs this prior distribution by incorporating the node embeddings to capture the semantic variations. 

\subsubsection{Contrastive-based Methods}

Contrastive-based methods enhance the discriminative power of learned representations by pulling positive samples closer and pushing negative samples farther apart to explore the semantic information. 
Under the graph contrastive learning framework, GDCL \cite{zhao2021graph} utilizes pseudo-labels to randomly select samples from classes different from positive samples to form negative samples, which corrects the sampling bias and thus decreases the false-negative samples in graph clustering. 
SCAGC \cite{xia2022self} further constructs the self-consistent contrastive loss by treating all samples from the same class in both graph views as positive samples and all non-matching samples as negative samples. 
CGC \cite{park2022cgc} utilizes a multi-level scheme for the selection of positive and negative samples, ensuring their ability to reflect hierarchical community structures and network homophily. Moreover, CGC extends its applicability to temporal graph clustering, which is capable of detecting change points.
Additionally, CONGREGATE \cite{sun2023congregate} reexamines graph clustering through a geometric lens. It constructs a novel heterogeneous curvature space for generating representations and introduces an augmentation-free reweighted contrastive method to focus more on both hard negatives and hard positives in the curvature space.
SCGC \cite{liu2023simple} incorporates a low-pass denoising operation in pre-processing, employs un-shared siamese encoders to eliminate the need for graph augmentation in contrastive learning, and introduces a cross-view structural consistency objective function to boost the discriminative capability of the learned network and avert direct clustering-guided loss.

\subsubsection{Discussion}

In addition to attribute graphs, numerous studies have explored clustering tasks on other graph types, such as heterogeneous graphs and dynamic graphs. 
Furthermore, graph-level clustering is another research-worthy issue~\cite{ju2023glcc}, but it has been relatively underexplored. Besides, graph clustering can be applied to practical applications, such as guiding recommendation services, analyzing protein-protein interaction networks, and uncovering cellular heterogeneity in single-cell RNA-Seq analyses, e.g. \cite{yu2023topological}.

\section{Semi-supervised Graph Learning}

\subsection{Classical Semi-supervised Graph Learning}

Semi-supervised learning is one of the most important tasks in machine learning, which manages to leverage an extensive corpus of unlabeled data to enhance the learning models trained using comparatively limited labeled examples. Compared with other semi-supervised learning methods, classical semi-supervised graph learning emphasizes the structural information of the graph. This focus enables a detailed exploration of the intrinsic relationships and dependencies among data points within the graph.

Based on the different data types within the graphs, classical semi-supervised graph learning can be categorized into node-level semi-supervised graph learning and graph-level semi-supervised graph learning. For the node-level task, the most representative methods are based on label propagation on the graph. For the graph-level task, the methods can be divided into consistency regularization and pseudo-labeling. 

\subsubsection{Label Propagation Methods}

Label propagation (LP) is the most representative method for label inference on semi-supervised graph learning. The framework can be formulated as a propagation process in which some of the nodes with labels, also referred to as seeds, propagate their labels to unlabeled nodes based on
the similarity of the connected nodes, which can be formulated as:
\begin{equation}
    \mathcal{L}=\mathcal{L}_{cls}(f(X), \mathcal{V}_L)+\lambda f(X)^{\top}\Delta f(X),
\end{equation}
where the first term $\mathcal{L}_{cls}$ is the classification loss which trains the model $f:\mathbb{R}^{N\times m}$ to predict the known labels $\mathcal{V}_L$, $m$ is the feature dimension. The second term is graph-based regularization, ensuring the connected nodes have a similar model output, $\Delta$ is the graph Laplacian, $\lambda\in\mathbb{R}$ is the regularization coefficient. For example, classical LP methods iteratively propagate the label of each data to its neighbors based on the constructed graph~\cite{zhu2005semi}. GCN~\cite{kipf2017semi} utilizes the power of graph neural networks and performs message passing on the graph to constrain neighborhood nodes with similar representations. GCN-LPA~\cite{wang2021combining} further proposes to combine GCN and LP with learnable edge weights. The model views LP as regularization to assist the GCN in learning proper edge weights. 


\subsubsection{Consistency Regularization Methods}

There are also some endeavors using consistency regularization for semi-supervised graph learning. These methods are based on the manifold or the smoothness assumption which posits that realistic perturbations of the graph data should not change the output of the model, and can be formulated as: 
\begin{equation}
\mathcal{L}=\mathcal{L}_s(f(X),\mathcal{G}_L)+\alpha\mathcal{L}_u(f(X))+\beta\mathcal{R}(f(X)),
\end{equation}
where $\mathcal{L}_s$ and $\mathcal{L}_u$ denote the supervised and unsupervised loss, $R$ denotes the consistency regularization loss. Typically, InfoGraph~\cite{sun2020infograph} learns a supervised and an unsupervised model respectively and maximizes the mutual information between the two models. GraphCL~\cite{you2020graph} and GLA~\cite{yue2022label} leverage contrastive learning between a graph and its augmented views to learn graph representation for semi-supervised graph learning. TGNN~\cite{ju2022tgnn} incorporates both graph neural network and graph kernels with consistency regularization loss to implicitly and explicitly explore graph structural knowledge from unlabeled data for semi-supervised graph classification.

\subsubsection{Pseudo-Labeling Methods}

Pseudo-labeling is another type of popular method for semi-supervised graph learning, which predicts the label distribution of unlabeled data and selects confident samples to the training data set as labeled data, which can be defined as:
\begin{equation}
\mathcal{L}=\mathcal{L}_s(f(X),\mathcal{G}_L)+\alpha\mathcal{L}_s(f(X'),\mathcal{G}'_U),
\end{equation}
where $X$ and $X'$ denote the supervised and selected unsupervised graph data feature, $\mathcal{G}_L$ and $\mathcal{G}'_U$ denote the label of the original labeled graph data and pseudo label of the unlabeled graph data respectively. 
For graph-structured data, SEAL-AI~\cite{li2019semi} and ASGN~\cite{hao2020asgn} leverage active learning techniques to select the most representative graph samples from the unlabeled data. KGNN~\cite{ju2022kgnn} adopts the posterior regularization to incorporate graph kernels as structured
constraints, generating pseudo labels and guiding the training process of graph neural networks
under the EM-style framework. DualGraph~\cite{luo2022dualgraph} further jointly learn a graph prediction and a graph retrieval module via posterior regularization during the pseudo-labeling process for semi-supervised graph learning.

\subsubsection{Discussion}
Label propagation methods represent nodes as training samples, and each edge denotes some similarity measurement of the node pair. In contrast, consistency regularization methods usually rely on the consistency constraint of rich
data transformations and pseudo-labeling methods rely on the high confidence of pseudo-labels, which can be added to the training data set as labeled data. 
Besides, some hybrid works integrating different types of methods into one unified framework~\cite{sohn2020fixmatch} can be adapted to graph data. 

\subsection{Semi-supervised Graph Learning under Domain Shift}

In real-world scenarios, graph applications often face out-of-distribution~(OOD) challenges, which arise when the data distribution during inference differs from the data on which the model was trained. Furthermore, the discrepancy between training and inference is complicated by the lack of labeled data in new domains, making supervised adaptation impractical.
To address these issues, the concept of Graph Domain Adaptation~(GDA) has been introduced to account for these distribution shifts and facilitate effective knowledge transfer.
To achieve data efficiency, GDA must operate with minimal reliance on large amounts of labeled data, which are often scarce or expensive to obtain. It emphasizes the value of leveraging prior knowledge and discriminative features from the available data.
Efficient GDA methods work to extend model adaptability with limited target domain data while maintaining robust performance despite rapid changes in data distribution.

Based on the data dependency, GDA can be categorized into unsupervised GDA, which does not require labeled data from the target domain, and source-free GDA, which performs adaptation without accessing the source domain data.

\subsubsection{Unsupervised Graph Domain Adaptation}

Unsupervised Graph Domain Adaptation~(UGDA) uses both source and target graphs for training, and tests primarily on the target graphs. 
UGDA methods can be broadly categorized into the following four types.

\paratitle{Discrepancy-based methods.} 
These methods usually apply discrepancy measurement, including MMD, Jensen–Shannon divergence, and Wasserstein distance, to measure the domain distribution shift. 
Formally, give source graphs $\mathcal{G}^{s}$, target graphs $\mathcal{G}^{t}$, and discrepancy measurement $d$, we have
\begin{equation}
    \mathcal{L}_{dis} = d(\mathcal{G}^{s}, \mathcal{G}^{t})\,,
\end{equation}
When the domain discrepancy loss $\mathcal{L}_{dis}$ is minimized, the knowledge can be transferred from the source domain to the target domain. 
Among them, UDAGCN~\cite{udagcn} exploits local and global consistency within graphs and integrates an attention mechanism to fuse these consistencies into node representations. 
GRADE~\cite{grade} introduces the graph subtree discrepancy measure to capture the distribution shift. 
SpecReg~\cite{SpecReg} proposes a tighter generalization bounded by spectral regularization. More recently, CoCo~\cite{coco} has achieved superior performance by synergizing two branches, including the graph message passing branch and the graph kernel branch.

\paratitle{Adversarial adaptation methods.}
These methods typically utilize adversarial components to minimize domain discrepancies, where the graph encoder and domain classifier compete.
Among them, AdaGCN~\cite{adagcn} employs adversarial domain adaptation techniques to acquire node representations that are invariant across varying domains, enabling effective knowledge transfer.
ACDNE~\cite{acdne} introduces adversarial learning across networks, in order to learn invariant node representations while preserving network structural information.
DANE~\cite{dane} is an adaptation method that attains domain-adaptive embedding through the utilization of a shared weight message passing network, complemented by adversarial learning regularization.

\paratitle{Discrimination learning methods.}
Researchers propose to utilize self-training to improve the performance in the target domain, where pseudo-labeling is one of the popular techniques. These methods obtain the pseudo-labels by the network itself on unlabeled target data to expand the training set. Among them, DEAL~\cite{deal} uses augmented views to distill reliable pseudo-labels for better graph-level classification in the target domain. StruRW~\cite{liu2023structural} introduces a structural reweighting approach that uses estimated edge probabilities based on the pseudo-labels to adjust the neighborhood bootstrapping of the graph neural network, counteracting conditional shifts between domains.

\paratitle{Disentangle-based methods.} 
These methods generally disentangle graph representation into domain-invariant and domain-relevant parts, and conduct domain adaptation with domain-invariant embeddings. 
Among them, ASN~\cite{asn} utilize both graph model embedding and adversarial adaptation to 
generate network-invariant node representations.
DGDA~\cite{DGDA} employs a variational graph auto-encoder approach to separate semantic, domain, and random latent variables for graph domain adaptation.

\subsubsection{Source-free Graph Domain Adaptation}

Source-free Graph Domain Adaptation~(SFGDA) eliminates the dependence on source data. The assumption that source and target data can be used for adaptation is not always true in real-world scenarios. On the one hand, privacy, confidentiality, and copyright issues may prevent access to the source data. On the other hand, the requirement to store the complete source dataset on devices is often impractical. 
Among SFGDA methods, GT3~\cite{gt3} utilizes a self-supervised test-time training framework, with a unique balancing constraint to prevent distribution bias.
GTrans~\cite{jin2023empowering} introduces a data-centric method for target data transformation to improve generalization and robustness.

\subsubsection{Discussion}
Graph domain adaptation aims at achieving data-efficient learning under domain shift. UGDA methods learn by modeling between domains, using labeled source domain data and unlabeled target domain data. SFGDA focuses on an even more data-efficient and more challenging problem, while the research in SFGDA is not comprehensive, and ongoing research continues to advance the field.

\section{Few-shot Graph Learning}
Few-shot graph learning aims to learn graph models to make accurate predictions with a small amount of labeled data, which usually adopts a meta-learning paradigm. Compared with semi-supervised graph learning concerning incomplete supervision, few-shot graph learning focuses on transferring prior knowledge across different tasks with high generalization capacity. Specifically, we aim to build a graph model that is readily tuneable to fit future meta-test tasks given a variety of meta-training tasks. In this section, we primarily concentrate on the most popular few-shot node classification and cover numerous few-shot graph classification works as well. Current few-shot graph learning approaches can be roughly divided into metric learning approaches and parameter optimization approaches, which we will elaborate on as below.

\subsection{Metric Learning Methods}
Metric learning methods typically incorporate ProNet \cite{snell2017prototypical} into graph few-shot learning, which generates prototypes by averaging the node representations of GNNs in the support set for each class. Then, they encourage each query node to approach its corresponding prototypes. Formally, given each graph $\mathcal{G}$ with $N$ classes and a query set $\mathcal{V}^q$ of $K$ query nodes for each class, we have 
\begin{equation}
\mathcal{L}_{ML}=-\sum_{(v_i,y_i) \in \mathcal{V}^q} \log \frac{\exp \left(-d\left(\mathbf{z}_{i}, \mathbf{p}_{y_i}\right)\right)}{\sum_{c} \exp \left(-d\left(\mathbf{z}_{i}, \mathbf{p}_{c}\right)\right)},
\end{equation}
where $\mathbf{p}_{c}$ denotes the representations for the $c$-th class, $\mathbf{z}_{i}$ is the node representation of $v_i$. $d(\cdot,\cdot)$ is a distance metric, e.g., cosine distance. The episodic training would then be executed iteratively until convergence is achieved. In particular, GFL~\cite{yao2020graph} integrates hierarchical graph representation gates to enhance node representations with structure-specific knowledge and employs a reconstruction loss to ensure stability during training. To improve the generalization capacity, GPN \cite{ding2020graph} estimates the importance of nodes for prototype reconstruction using a second GNN with additional adjustments based on centrality. HAG-Meta \cite{tan2022graph} additionally takes into account the scenarios involving newly encountered classes and introduces a regularization term based on task-level and novel-level attention to tackle the potential class imbalance issue. 
HGNN~\cite{yu2022hybrid} combines prototype GNN with an instance GNN, which consists of all support nodes and the query, and encourages the consistency between these complementary GNNs. 
SuperClass~\cite{chauhan2020few} further computes prototype graphs for few-shot graph classification by minimizing the average distance to other similar graphs. Subsequently, it constructs a graph-of-graph to represent inter-class correlations for generalized graph learning.

\subsection{Parameter Optimization Methods}
Parameter optimization methods usually adhere to the paradigm of model-agnostic meta-learning (MAML)~\cite{finn2017model,liu2020does}, which generates node representations using GNNs for meta-learning. They construct meta-tasks using random nodes and then conduct multiple gradient descent for each meta-task. Formally, for each sampled meta-task $\mathcal{T}_{i}$, given a GNN $f_\theta(\cdot)$ 
and learning rate $\lambda$, we have:
\begin{equation}
\theta_{i}^{\prime}=\theta-\lambda \frac{\partial \mathcal{L}_{\mathcal{T}_{i}}\left(f_{\theta}\right)}{\partial \theta}.
\end{equation}
Finally, these tasks would be aggregated to optimize the task-agnostic parameters as a whole:
\begin{equation}
\theta \leftarrow \theta-\beta \nabla_{\theta} \sum_{\mathcal{T}_{i} \sim p(\mathcal{T})} \mathcal{L}_{\mathcal{T}_{i}}\left(f_{\theta_{i}^{\prime}}\right),
\end{equation}
where $\beta$ is the learning rate for meta-optimization. 
On the basis of the above optimization framework, numerous GNN approaches are developed. In particular, Meta-GNN~\cite{zhou2019meta} is the first work to combine meta-learning with GNNs and validate the effectiveness of GNNs in few-shot learning scenarios. 
AMM-GNN~\cite{wang2020graph} further incorporates the attention mechanism to capture the relationships among meta-tasks for more generalized knowledge. GLITTER~\cite{wang2022graph} extracts relevant nodes according to their influence to get rid of potential interference and then constructs a task-specific graph by leveraging mutual information and nodal influence, which makes the best of meta-tasks for effective meta-learning.  
In regard to few-shot graph classification, AS-MAML~\cite{ma2020adaptive} analyzes the difficulty of applying MAML in graph domains and thus introduces an adaptive step controller that takes into account both training loss and the quality of graph embeddings to regulate the optimization procedure. HSL-RG~\cite{ju2023few} investigates the correlation between graph samples using graph kernels from a global perspective
and introduces various augmentation strategies for self-supervised learning from a local perspective. The attention mechanism further integrates both complementary views into a task-adaptive meta-learning framework. 


\subsection{Discussion}
In addition to the above works, several advanced techniques including prompt learning and contrastive learning have also been applied in few-shot graph learning~\cite{tan2023virtual}. 
Besides, these few-shot learning works have also been extended to tackle zero-shot graph learning~\cite{ju2023zero}, which focuses on nodes from unseen classes. 
\section{Conclusion and Future Work}
In summary, this paper provides a comprehensive overview of Data-Efficient Graph Learning. We initiate by discussing the current challenges in graph machine learning. Subsequently, we categorize existing works into three parts: self-supervised graph learning, semi-supervised graph learning, and few-shot graph learning. In each section, we present representative strategies and introduce their key technical contributions. Despite the progress made, there still remain several challenges in the field that warrant further research in the future.

\paragraph{Inherent Challenges.}
Though being efficient in general, the data-efficient graph learning models are inherently facing challenges such as robustness and generalizability.
We might also face extra challenges when running data-efficient models on out-of-distribution data. 
Some real-world applications prefer explainable results, so that the model can be trusted. Drug discovery is a good example. 
All these considerations not only enhance the capabilities of existing data-efficient graph learning models but also steer the trajectory of research towards more versatile and reliable solutions for real-world scenarios.

\paragraph{Combined with Large Models.}
Recently, combining LLM (i.e., Large Language Models) with graph learning approaches is one of the trending topics~\cite{yang2023poisoning}. 
For example, one might use GNN components to learn the underlying knowledge graphs to improve LLMs for question-answering~\cite{yasunaga2021qa} or use LLMs to generate node embeddings in social network graphs~\cite{liu2021content}.
For data-efficient graph learning, we care about all components. Both the LLM-based components and GNN-based components are crucial for achieving effectiveness and efficiency. It is theoretically possible to achieve high performance with only few-shot or zero-shot learning on these relatively-complicated models.

\paragraph{Towards Different Convolutional Algorithms.}
There are some GNN models that are fundamentally different from most of the others. Therefore, data-efficient learning approaches that work elsewhere might no longer be applicable. For example, some GNNs are designed in non-Euclidean space. 
The non-Euclidean-space models (e.g., spherical or hyperbolic) are well-known for being good at handling certain substructures, such as trees or cycles.
Researchers are still on their way to extend GNN models to non-Euclidean spaces, while believing it to be a promising direction.
The data-efficient design in Euclidean space might no longer work for non-Euclidean space, and there might be data-efficient non-Euclidean GNN models that can not work in Euclidean space, highlighting the continuous evolution and diversification of GNN research.

\paragraph{Proof of Efficiency.}
In theory, data-efficient learning has the potential to be rigorously justified by mathematical proofs. So far, with the help of learnability theory, researchers have made strides in proving some theoretical bound under certain problem settings~\cite{grohs2021proof}. 
However, so far, the majority of works in this field focus on fully-connected neural networks, together with fully-labeled data. If mathematical analysis could be provided in the data-efficient graph learning setting, it would undoubtedly provide valuable insights for future researchers, paving the way for more robust and informed advancements in this evolving domain.

\section*{Acknowledgments}
This paper is partially supported by the National Natural Science Foundation of China (NSFC Grant Numbers 62306014 and 62276002) as well as the China Postdoctoral Science Foundation with Grant No. 2023M730057.

\bibliographystyle{named}
\bibliography{ijcai24}

\end{document}